\documentclass[
]{ceurart}

\usepackage{listings}
\usepackage{xcolor}
\usepackage[most]{tcolorbox}

\begin{document}

\copyrightyear{2025}
\copyrightclause{Copyright for this paper by its authors.
  Use permitted under Creative Commons License Attribution 4.0
  International (CC BY 4.0).}

\conference{SESAME 2025: Smarter Extraction of ScholArly MEtadata using Knowledge Graphs and Language Models, @ JCDL 2025}

\title{How Do LLMs Encode Scientific Quality? An Empirical Study Using Monosemantic Features from Sparse Autoencoders}



\author[1]{Michael McCoubrey}[%
orcid=0009-0002-8628-9417,
email=michael.mccoubrey@open.ac.uk
]
\cormark[1]
\address[1]{Knowledge Media Institute, The Open University,
Walton Hall,
Milton Keynes,
MK7 6AA,
United Kingdom}

\author[1]{Angelo Salatino}[%
orcid= 0000-0002-4763-3943,
email=angelo.salatino@open.ac.uk,
]

\author[1,2]{Francesco Osborne}[%
orcid=0000-0001-6557-3131,
email=francesco.osborne@open.ac.uk,
]
\address[2]{Department of Business and Law, University of Milano Bicocca, Milan, IT}

\author[1]{Enrico Motta}[%
orcid=0000-0003-0015-1952,
email=enrico.motta@open.ac.uk,
]

\cortext[1]{Corresponding author.}


\begin{abstract}
In recent years, there has been a growing use of generative AI, and large language models (LLMs) in particular, to support both the assessment and generation of scientific work. Although some studies have shown that LLMs can, to a certain extent, evaluate research according to perceived quality, our understanding of the internal mechanisms that enable this capability remains limited. 
This paper presents the first study that investigates how LLMs encode the concept of scientific quality through relevant monosemantic features extracted using sparse autoencoders. We derive such features under different experimental settings and assess their ability to serve as predictors across three tasks related to research quality: predicting citation count, journal SJR, and journal h-index. The results indicate that LLMs encode features associated with multiple dimensions of scientific quality. In particular, we identify four recurring types of features that capture key aspects of how research quality is represented: 
1) features reflecting research methodologies; 2) features related to publication type, with literature reviews typically exhibiting higher impact; 3) features associated with high-impact research fields and technologies; and 4) features corresponding to specific scientific jargons.
These findings represent an important step toward understanding how LLMs encapsulate concepts related to research quality.
\end{abstract}



\maketitle

\section{Introduction}

In recent years, the research evaluation community has increasingly investigated the potential of generative AI to support the assessment of scientific work~\cite{andersen2025generative,bolanos2024artificial}. Several initiatives have emerged that employ large language models (LLMs) to evaluate, improve, or even generate research papers and other scholarly documents~\cite{thelwall2025evaluating,brody2021scite,buscaldi2024citation,meloni2023integrating}. These models have been applied, for example, to assist the peer review process for both papers~\cite{thelwall2025evaluating} and grant proposals~\cite{cortes2024ai}, and their adoption is expected to grow further. LLMs are also increasingly used to generate literature reviews~\cite{bolanos2024artificial,osborne2019reducing}, demonstrating a strong ability to synthesise and summarise scientific knowledge, although the quality of current systems remains below optimal standards. Furthermore, recent studies have investigated how LLMs and related AI technologies can support the generation of novel research hypotheses~\cite{alkan2025survey,borrego2025research}, as well as the synthesis of scientific knowledge into structured representations such as knowledge graphs~\cite{dessi2025cs,john2025scimantify,tsaneva2025knowledge} and scientific taxonomies~\cite{salatino2025survey,aggarwal2026large}.
These advances suggest that such technologies may become integral components of the scientific discovery process.
Beyond these research efforts, numerous studies have shown that LLMs are increasingly used in scientific writing, raising important ethical questions regarding authorship, bias, and transparency~\cite{andersen2025generative, lehmann2024large}. These issues are becoming more pressing as the performance of LLMs continues to improve and their influence on research practice expands.

Given the growing importance of LLMs as tools for scientific research and the production of scholarly materials, it is crucial to develop a deeper understanding of how they represent and encode concepts relevant to this domain, particularly the notion of scientific quality. While several studies~\cite{thelwall2025evaluating,cortes2024ai,andersen2025generative} have shown that LLMs can, to some extent, evaluate research papers according to perceived quality, there remains a lack of research on the internal mechanisms that enable this capability. This gap is crucial, considering the central role of quality assessment in research evaluation.

Progress in this area has been limited partly because LLMs largely function as black boxes, whose internal reasoning processes are difficult to interpret. However, recent advances in LLM explainability have introduced promising approaches based on the concept of monosemantic features~\cite{paulo2024automatically}, which can offer new insights into how these models process and evaluate scientific information. Monosemantic features, typically identified using sparse autoencoders~\cite{shu2025survey}, are internal model representations that correspond to distinct and interpretable concepts, such as specific topics or entities. 

This paper presents the first systematic investigation into how LLMs encode the concept of \textit{scientific quality} through relevant monosemantic features. We address this question by identifying and analysing monosemantic features that are predictive of research quality across multiple models. The insights derived from this study represent an initial step toward explaining how the internal representations of LLMs capture this elusive yet fundamental notion, and how these features may contribute to a more transparent understanding of the ability of LLMs to assess and generate scientific research.

To identify the monosemantic features linked with research quality, we first extract such features using sparse autoencoders across three different configurations that employ Gemma 2 2B and Gemma 2 9B in combination with various autoencoders. The extracted features are then used to train a decision tree classifier that predicts the relative quality of research papers according to three bibliometric proxy indicators: 1) the citation count quartile of each paper, 2) the journal SJR quartile of the venue in which the paper was published, and 3) the journal h-index quartile of the same venue.
The trained classifiers achieved fair performance across all tasks, indicating that the selected features are predictive of research quality, at least within the context of these proxies. We then analyse which specific features are most informative for each task and examine their presence across the three configurations. 
All experimental materials are available in the associated repository\footnote{Repository - \url{https://github.com/McCoubreym/How-Do-LLMs-Encode-Scientific-Quality-}}. 

The results indicate that LLMs encode features associated with multiple dimensions of research quality. In particular, we identify four recurring types of features that emerge across dimensions and capture key aspects of how research quality is represented in textual form. These features suggest that LLMs internalise latent cues that humans also use, consciously or unconsciously, to assess the rigour, relevance, and impact of scientific work.

The first feature type relates to \textit{research methodology}. LLMs appear to associate higher quality with texts that describe rigorous methodological approaches, whether qualitative or quantitative, and that explicitly mention the use of established research protocols or well-defined experimental procedures. 
The second type concerns the \textit{publication type}. The models seem to recognise that survey papers and literature reviews tend to be of higher scholarly impact, possibly because they provide comprehensive overviews, methodological syntheses, or conceptual frameworks that guide subsequent research. 
The third type involves \textit{research fields and technologies}. References to emerging, high-impact, or rapidly developing domains, subfields, or technologies appear to act as strong signals of perceived research importance. This suggests that LLMs capture topical trends and the social dynamics of research communities. 
Finally, we observe a less interpretable but intriguing group of features related to \textit{academic language and specialised scientific jargon}. These patterns often correspond to community-specific terminologies and writing conventions, which may implicitly encode information about disciplinary identity and methodological preferences. A possible interpretation is that LLMs learn stylistic and lexical regularities characteristic of specific research communities. However, the precise role of this feature group remains unclear and warrants further investigation.

\section{Related Work}\label{sec:relatedwork}

In recent years, several initiatives have investigated the use of LLMs for evaluating scientific research and generating research-related content. These studies explore whether LLMs can emulate the analytical and evaluative capabilities of human experts.

A few studies have investigated the ability of language models to generate peer review scores and reports~\cite{thelwall2025evaluating}. Although recent models can produce evaluations that are broadly comparable to those of human reviewers, challenges related to consistency, transparency, and bias remain unresolved. Related research has also explored the use of LLMs in grant peer review processes within funding organisations, such as the \textit{la Caixa} Foundation~\cite{cortes2024ai}.  
Another emerging application is the automatic generation of related work sections using retrieval-augmented generation (RAG) pipelines~\cite{fan2024survey}, which are employed to produce summaries of the  literature~\cite{bolanos2024artificial}. LLM-based systems have also been applied to reference recommendation~\cite{buscaldi2024citation,brody2021scite}, the classification of research papers~\cite{xu2020building,cadeddu2024comparative,salatino2022cso} and scientific sentences~\cite{bolanos2025modelling}, hypothesis generation~\cite{alkan2025survey,borrego2025research}, the construction of structured representations of scientific knowledge~\cite{john2025scimantify,dessi2025cs}, and the development of conversational agents for supporting literature analysis~\cite{meloni2023integrating,laranjo2018conversational}.
These applications implicitly rely on the model’s capacity to identify rigorous and relevant research, suggesting that LLMs encode an internal notion of \textit{research quality}. However, this capability remains poorly understood due to the opaque nature of current models. 

To investigate how LLMs encode and reason about abstract concepts, recent research has introduced the notion of monosemantic features~\cite{paulo2024automatically}. Unlike traditional embeddings that conflate multiple meanings, monosemantic features aim to disentangle distinct semantic dimensions into interpretable, context-independent components. They are typically identified using Sparse Autoencoders (SAEs)~\cite{shu2025survey}, which represent model activations through a sparse set of interpretable features. 
Recent studies have shown that this approach can uncover fine-grained, conceptually coherent structures often missed by conventional interpretability methods~\cite{bricken2023monosemanticity, templeton2024scaling}. 

In this paper, we extend the state of the art by being the first to investigate how LLMs represent the concept of research quality using the monosemantic feature methodology. 

\section{Methodology}\label{sec:methodology}

In this section, we describe the methodology employed to identify monosemantic features that are predictive of high-quality research. The process comprised four main stages: 1) \textbf{data preprocessing}, which involved selecting and preparing the dataset; 2) \textbf{feature extraction}, performed by using LLMs and SAEs to generate research paper summaries and extract monosemantic features; 3) \textbf{model training}, which consisted of training decision tree classifiers on the extracted features across the three binary classification tasks; and 4) \textbf{qualitative analysis}, in which we examined the decision tree features to identify the monosemantic concepts associated with research quality.

The dataset used in this experiment was derived from the Academia/Industry DynAmics Knowledge Graph (AIDA KG)~\citep{angioni2021aida}, a large-scale knowledge base comprising approximately 25 million Computer Science publications. The papers in this knowledge graph are described by a rich set of metadata, including authors, organisations, countries, venues, and associated research topics.  
For this study, we extracted a corpus of 38,639 papers and characterised each publication by its title, abstract, five-year citation count~\cite{abramo2011assessing}, and publishing journal. 
To provide additional context on journal quality, we enriched the dataset with two journal-level indicators: the SJR value
 and the h-index, both obtained from the Scimago Journal \& Country Rank (SJR). 
The SJR value measures the scientific influence of a journal by considering both the number of citations it receives and the prestige of the citing journals~\cite{guerrero2012further}. The h-index reflects the journal’s productivity and citation impact, corresponding to the maximum number \( h \) such that the journal has published \( h \) articles that have each received at least \( h \) citations~\cite{hodge2011evaluating}. 

Although the publication metrics used in this analysis are imperfect proxies for research quality, they are utilised due to their widespread availability.

We then converted the three metrics of interest (citation count, SJR value, and h-index) into quartiles to capture relative performance levels across papers and journals. To focus on the most distinctive cases, we selected only the first (highest) and fourth (lowest) quartiles of each metric and constructed three balanced subsets for binary classification. These subsets are half the size of the full corpus of 38,639 papers. 
These subsets were designed to train classifiers for three specific prediction tasks regarding \textit{Citation Count Quartiles}, \textit{SJR Value Quartiles}, and \textit{Journal h-index Quartiles}. Each dataset was split into training and test sets, following a 70:30 ratio.


Next, we extracted monosemantic features to serve as representations of the papers in the classification tasks. To this end, we explored different configurations to evaluate whether the extracted features remain consistent across various models and SAEs. 
Specifically, we employed the following settings: 1) Gemma 2 2B with the SAE \textit{gemma-scope-2b-pt-res} (16k features version) at layer 20; 2) Gemma 2 9B-it with the SAE \textit{gemma-scope-9b-it-res} (16k features version) at layer 20; and 3) Gemma 2 9B-it with the SAE \textit{gemma-scope-9b-it-res} (16k features version) at layer 31 \cite{lieberum2024gemmascopeopensparse}\cite{gemmateam2024gemma2improvingopen}. 
To extract the monosemantic features, we prompted each LLM to generate a three-sentence summary for every research paper. 
The prompts used for this task are provided in the repository. 
Subsequently, we computed the features for each token in the generated summaries using the SAE, and then averaged these features across all tokens within each summary. This procedure produced a single, interpretable vector for every research paper. A three-sentence summary was chosen to standardize the number of tokens being averaged, preventing length-based distortion in the paper-level features.

The resulting features were used as paper representations to train a decision tree classifier for the three binary classification tasks defined earlier. This classifier was selected because it 1) offers a highly interpretable representation of the decision-making process based on the input features, and 2)  performs feature subset selection, ensuring that the resulting explanations depend on a limited set of features. The maximum number of leaf nodes was optimised for each task by evaluating model performance through cross-validation on the LLM \& SAE combination 2. The optimised values for each task were then applied to the other settings to ensure comparability across decision trees. 
Finally, a standard BERT classifier~\cite{devlin2019bert} was trained to predict the quartiles of the research quality metric using a tokenised version of the same prompt adopted in the first setting. This experiment aimed to benchmark the performance of the decision trees against that of a state-of-the-art text classification model.

Following the generation of decision trees for each task, we performed a qualitative analysis to interpret the model’s predictive logic.
To determine the conceptual meaning of each monosemantic feature in the resulting trees, we employed the Neuropedia service~\cite{neuronpedia}, which supports the investigation of feature behaviour by comparing texts with different activation values, identifying the tokens that maximally activate a feature, and observing how an LLM’s generated text changes when a feature’s activation value is artificially clamped to a fixed level.
To further enhance the quality of the feature definitions, we examined the titles and abstracts of papers that displayed substantially different activation values for a given monosemantic feature of interest. 
After determining the meaning of individual features, our analysis aimed to group them into common themes to facilitate their interpretation. 

\section{Results and Discussion}\label{sec:results}

\begin{table}[]
\caption{Accuracy of a decision tree classifier trained on monosemantic features extracted from different combinations of LLMs and SAEs across Tasks 1–3.}\label{tab:accuracy}
\scriptsize
\begin{tabular}{r|c|c|c|c|c}
\toprule
           & \textbf{LLM+SAE, 1 comb.} & \textbf{LLM+SAE, 2 comb.} & \textbf{LLM+SAE, 3 comb.} & \textbf{Task   Avg} & \textbf{BERT   Classifier} \\
           \midrule
Task 1     & 0.6666           & 0.7211           & 0.6486           & 0.6788     & 0.6522            \\
Task 2     & 0.7593           & 0.7463           & 0.7328           & 0.7461     & 0.8082            \\
Task 3     & 0.7088           & 0.7351           & 0.7490            & 0.7310      & 0.8274            \\
Avg. Model & 0.7116           & 0.7342           & 0.7101           &    0.7186        & 0.7626            \\
\bottomrule
\end{tabular}
\end{table}

In this section, we first discuss the performance of the classifier and then analyse the most significant features for each task. 
Table~\ref{tab:accuracy} reports the classification accuracy of the classifiers across the three tasks. Because the dataset is perfectly balanced, accuracy was chosen as the evaluation metric. It offers the most intuitive measure of performance without the bias often found in imbalanced datasets, where F1-scores are typically required.
It should be emphasised that the purpose of this study is not to achieve state-of-the-art classification performance. Instead, the classifiers are employed as analytical instruments to examine the information content and interpretability of the monosemantic features extracted from LLMs. 

Interestingly, the decision tree classifier, which relies exclusively on the monosemantic features, performs slightly better than the BERT-based baseline in predicting the citation count quartile. This finding indicates that the monosemantic features capture meaningful and discriminative information relevant to this task. In contrast, the BERT classifier attains higher accuracy on the other two tasks, as expected for a model directly optimized on textual representations rather than interpretable features.

Considering the decision tree classifiers, the highest predictive performance was achieved for Task 2 (SJR quartile prediction), followed by Task 3 (journal h-index quartile prediction) and Task 1 (citation count quartile prediction). Overall, these results indicate that the monosemantic features extracted from the LLMs exhibit substantial predictive power with respect to the three metrics used as proxies for research quality.


The predictive performance remained broadly consistent across different combinations of LLMs and SAEs. The largest LLM (9B), when paired with an SAE at layer 20, achieved a slightly higher accuracy than the other configurations. This result suggests a modest benefit in using larger models and extracting monosemantic features from middle-layer activations.

The most influential features identified by the classifier varied across the three tasks. This variation indicates that the quality metrics capture distinct dimensions of research impact or possibly reflect different underlying phenomena. 
The citation quartile prediction task was distinctive, as the resulting decision tree structure was consistently reproduced across different LLM and SAE combinations. This finding suggests that the concepts underlying a paper’s citation count are more stable and coherent across models. 
In the following subsection, we provide a detailed discussion of the results obtained for the three tasks. Due to space constraints, we report only the main features, while the full classification trees are available in the appendix.

\subsection{Task 1 - Predicting citation quartiles}

\begin{table}[]
\caption{The monosemantic features for Task 1 (Predicting citation quartiles) across the 3 settings. Rows are ordered by setting number and feature importance. Blue index numbers indicate a positive association with the proxy metric, while orange ones indicate a negative association. More details about the decision trees are provided in Figures \ref{fig:t1_2b_20}, \ref{fig:t1_9b_20}, and \ref{fig:t1_9b_31} in the Appendix.}
\label{task_1_features}
\scriptsize
\begin{tabular}{r|r|l}
\toprule
\textbf{LLM \& SAE} & \textbf{Index}                        & \textbf{Description}                                                                                  \\ 
\midrule
1          & {\color[HTML]{ef8a62} 74}    & Overfocus on methodological details.              \\ 
1          & {\color[HTML]{67a9cf} 15676} & Academic survey and reviews.              \\ 
1          & {\color[HTML]{ef8a62} 6631}  & Has an ambiguous definition. Likely linked to the length of the text.                        \\ 
1          & {\color[HTML]{67a9cf} 2297}  & Names, particularly about technologies but can include people and institutions.              \\ 
1          & {\color[HTML]{67a9cf} 12691} & Terms and phrases used in physics.                                                           \\ 
2          & {\color[HTML]{67a9cf} 6135}  & Names of popular technologies e.g. big data, cloud computing, internet of things etc. \\ 
2          & {\color[HTML]{67a9cf} 979}   & Academic survey and reviews.                                                                 \\ 
2          & {\color[HTML]{ef8a62} 15531} & Has an ambiguous definition. Potentially about academic style of language.                   \\ 
2          & {\color[HTML]{67a9cf} 13161} & Tools related to inventions and innovations.                                                 \\ 
2          & {\color[HTML]{ef8a62} 61}    & Has an ambiguous definition. Potentially about academic style of language.                   \\ 
3          & {\color[HTML]{67a9cf} 12640} & Names, particularly about technologies but can include people and institutions.              \\ 
3          & {\color[HTML]{ef8a62} 6355}  & Has an ambiguous definition. Potentially a methodology-focused paper.                                                                            \\ 
3          & {\color[HTML]{67a9cf} 2356}  & Academic survey and reviews.                                                                 \\ 
3          & {\color[HTML]{ef8a62} 6427}  & Has an ambiguous definition. Potentially related to the methods used.                        \\ 
3          & {\color[HTML]{67a9cf} 3156}  & Names, particularly about technologies.                                                      \\ 
\bottomrule
\end{tabular}
\end{table}

The monosemantic features employed by the decision trees for task 1 are reported in Table \ref{task_1_features}. These features can be grouped into three main categories.

The first category concerns the \textbf{type of paper and the research methodology} (e.g., 979, 2356, 6355, 15676), that is, whether the publication is a survey, a literature review, or an empirical study. As expected, surveys and literature reviews tend to be associated with a higher citation count.

The second category concerns the presence of \textbf{specific technologies} (e.g., 2297, 13161, 3156, 12640). For instance, we identify features such as 6135 and 13161, which are directly associated with several popular technologies (big data, cloud computing, internet of things, Hadoop).
This category is relatively easy to interpret, as publishing on emerging technologies at the right time often results in higher citation rates. Furthermore, previous studies have shown~\cite{corrin2022importance} that the use of such specific, topical keywords in the title and abstract is a key factor in Academic Search Engine Optimisation, which in turn improves the discoverability of a publication and can increase its citation count.

These first two categories of features indicate that the main themes emerging from the model are closely linked to a publication’s discoverability, which in turn shapes its citation impact. Discoverability depends on the publication’s nature and topical focus, as these influence how easily other researchers can locate and reuse its content. This interpretation accords with Information Foraging Theory~\cite{pirolli1999information}, which views scientists as foragers seeking to maximise informational value while minimising search effort and uncertainty. Publications that bridge structural gaps or introduce reusable technologies strengthen the “information scent” of the literature, making them more visible and useful to others. As a result, they tend to attract citations rapidly, reflecting the interplay between information foraging and cumulative advantage in citation dynamics.

The third and more difficult category to interpret involves the use of \textbf{academic language} (e.g., 74, 61, 15531), which captures the extent to which a publication employs specialised scientific jargon. These may suggest that multiple subtle linguistic aspects may be linked with citation rates. However, the specific mechanisms behind this relationship are not yet clear. These features, therefore, warrant further investigation in future work.

\begin{table}[]
\caption{The monosemantic features for Task~2 (Predicting SJR quartiles) across the 3 settings. The order and colours follow the same convention as in the previous table. More details about the decision trees are provided in Figures~\ref{fig:t2_2b_20}, \ref{fig:t2_9b_20}, and \ref{fig:t2_9b_31} in the Appendix.}
\label{task_2_features}
\scriptsize
\begin{tabular}{r|r|l}
\toprule
\textbf{LLM \& SAE} & \textbf{Index}                        & \textbf{Description}                                                                                        \\ 
\midrule
1          & {\color[HTML]{67a9cf} 7402}  & Reference to digital transmission codes.                                                           \\ 
1          & {\color[HTML]{67a9cf} 16116} & Refers to feature (particularly feature detection) used in machine   learning.                     \\ 
1          & {\color[HTML]{67a9cf} 8411}  & Has an ambiguous definition. References to mobile phones, wireless   networks and sensors.         \\ 
1          & {\color[HTML]{67a9cf} 12691} & Academic style language.                                                                           \\ 
1          & {\color[HTML]{67a9cf} 9954}  & Vehicle specifications.                                                                            \\ 
1          & {\color[HTML]{ef8a62} 4897}  & Has an ambiguous definition. Terms related to biological and physical   activities.                \\ 
1          & {\color[HTML]{67a9cf} 12679} & Has an ambiguous definition. The use of punctuation and formatting within   sentences.             \\ 
1          & {\color[HTML]{67a9cf} 26}    & References to leadership, management, organisations.                                               \\ 
1          & {\color[HTML]{67a9cf} 15324} & Standards for mobile communication e.g. GSM, LTE, LTE Advanced.                                    \\ 
1          & {\color[HTML]{67a9cf} 10026} & Reference to computational complexity theory.                                                      \\ 
1          & {\color[HTML]{67a9cf} 4785}  & Referring to a scientific research publication.                                                    \\ 
1          & {\color[HTML]{67a9cf} 12938} & Refers to electrical components and systems, particularly about the   electric grid.               \\ 
2          & {\color[HTML]{67a9cf} 16326} & Physics and engineering terms and concepts.                                                        \\ 
2          & {\color[HTML]{67a9cf} 1582}  & Wireless communication technologies.                                                               \\ 
2          & {\color[HTML]{67a9cf} 599}   & Computer vision.                                                                                   \\ 
2          & {\color[HTML]{67a9cf} 14018} & Terms related to ML feature extraction methods.                                                    \\ 
2          & {\color[HTML]{ef8a62} 15358} & References to academic publications and presenting academic work                                   \\ 
2          & {\color[HTML]{ef8a62} 4062}  & Words commonly within code and software development.                                               \\ 
2          & {\color[HTML]{ef8a62} 15378} & Clinical trial and randomised controlled trials.                                                      \\ 
2          & {\color[HTML]{67a9cf} 7671}  & Training CNN for computer vision and computer vision training features.                            \\ 
2          & {\color[HTML]{67a9cf} 1063}  & Refers to correlations in data.                                                                    \\ 
2          & {\color[HTML]{67a9cf} 674}   & Has an ambiguous definition.                                                                       \\ 
3          & {\color[HTML]{67a9cf} 10403} & Terms associated with data transmission particularly signal modulation.                            \\ 
3          & {\color[HTML]{67a9cf} 12285} & Image based machine learning e.g. denoising, reconstruction,   decomposition, super-resolution.    \\ 
3          & {\color[HTML]{67a9cf} 9816}  & Standards for mobile communication e.g. GSM, LTE, LTE Advanced.                                    \\ 
3          & {\color[HTML]{67a9cf} 1715}  & Has an ambiguous definition. Related to algorithms and methods for data   processing and analysis. \\ 
3          & {\color[HTML]{67a9cf} 15246} & Machine learning techniques.                                                                       \\ 
3          & {\color[HTML]{67a9cf} 15265} & Words with the prefix ’it’ . Mostly referring to information technology   e.g. IT governance.      \\ 
3          & {\color[HTML]{67a9cf} 5001}  & Electricity grid technologies and management.                                                       \\ 
3          & {\color[HTML]{67a9cf} 1436}  & Camera and image tasks, particularly camera calibration.                                           \\ 
3          & {\color[HTML]{ef8a62} 13675} & Phrases related to copyright and licensing information.                                            \\ 
3          & {\color[HTML]{67a9cf} 15237} & References to leadership, management, organisations.                                               \\ 
3          & {\color[HTML]{67a9cf} 15301} & Statistical terminology.                                                                           \\ 
\bottomrule
\end{tabular}
\end{table}

\subsection{Task 2 - Predicting SJR quartiles}
The features that emerged as the best predictors for classifying papers according to the SJR quartile of their publication venue (reported in Table \ref{task_2_features}) differ substantially from those identified for predicting citation counts. 

The first and most prominent group of features relates to \textbf{research fields and field-specific technologies}. These features typically indicate the disciplinary or technological focus of a publication and often reveal the subdomain in which the research is situated. Many of these field-related features were consistently retrieved across all three settings, confirming their strong relevance to this task. In particular, features associated with communication networks, electricity grids, data transmission, and business emerged as key indicators. 
The prevalence of these features suggests that certain domains are systematically associated with higher SJR quartiles. This can be partly explained by the structure of the SJR index~\cite{guerrero2012further}, which, while normalising for subject area, still attributes greater weight to citations from more prestigious journals. As a result, papers in research areas where highly ranked journals frequently cite one another tend to be linked to venues with higher SJR values. This residual disciplinary bias creates a strong connection between the thematic content of a paper and the prestige level of its publishing journal.

A second important group of features concerns the \textbf{research methodologies} (e.g., 15378, 1715, 1063) adopted by the publications. These include explicit references to experimental protocols, such as randomised controlled trials, as well as methodological approaches based on machine learning and standard data processing techniques. Such features capture the analytical rigour and methodological design of the research rather than its topical focus.
These features, which capture both the structural soundness and credibility of a study, appear to correlate with publication in higher-quality journals, suggesting that methodological rigor serves as a key indicator of perceived research quality. Their presence in the embeddings learned by the LLMs indicates an implicit understanding of the characteristics that define high-quality research.

\begin{table}[]
\caption{The monosemantic features for Task~3 (Predicting journal h-index quartiles) across the 3 settings. The order and colours follow the same convention as in the previous table. More details about the decision trees are provided in Figures \ref{fig:t3_2b_20}, \ref{fig:t3_9b_20}, and \ref{fig:t3_9b_31} in the Appendix.}
\label{task_3_features}
\scriptsize
\begin{tabular}{r|r|l}
\toprule
\textbf{LLM \& SAE} & \textbf{Index}                        & \textbf{Description}                                                                              \\ \midrule
1          & {\color[HTML]{ef8a62} 9890}  & Educational methodology and technology.                                                  \\ 
1          & {\color[HTML]{67a9cf} 6278}  & Acknowledging contributions and support from others in research.                         \\
1          & {\color[HTML]{ef8a62} 1881}  & Software development language particularly about software architectures.                 \\ 
1          & {\color[HTML]{67a9cf} 15324} & Standards for mobile communication e.g. GSM, LTE, LTE Advanced.                          \\ 
1          & {\color[HTML]{67a9cf} 12691} & Complex academic language.                                                               \\ 
1          & {\color[HTML]{67a9cf} 6465}  & Fuzzy logic/sets/values.                                                                 \\ 
2          & {\color[HTML]{ef8a62} 1392}  & Qualitative research methodologies.                                                      \\ 
2          & {\color[HTML]{67a9cf} 1618}  & Data analysis techniques.                                                                \\ 
2          & {\color[HTML]{ef8a62} 3864}  & Complex academic language.                                                               \\ 
2          & {\color[HTML]{67a9cf} 5788}  & Technical terms and concepts related to telecommunications and   networking.             \\ 
2          & {\color[HTML]{67a9cf} 3919}  & Contact information and author-related details.                                          \\ 
2          & {\color[HTML]{ef8a62} 15358} & Common academic expressions.                                                             \\ 
3          & {\color[HTML]{ef8a62} 8721}  & Discussion, exploring, or providing an overview/summary. Linked to   literature reviews. \\ 
3          & {\color[HTML]{67a9cf} 10557} & Telecommunication technologies.                                                          \\ 
3          & {\color[HTML]{ef8a62} 15809} & Computing architectures and algorithms.                                                  \\ 
3          & {\color[HTML]{67a9cf} 6980}  & Requests for or references to contact information.                                       \\ 
3          & {\color[HTML]{ef8a62} 2543}  & Pedagogic interventions.                                                                 \\ \bottomrule
\end{tabular}
\end{table}

\subsection{Task 3 - Predicting journal h-index quartiles}
The monosemantic features that emerged as strong predictors for classifying papers according to the h-index quartile of their publication venues (Table \ref{task_3_features}) combine elements identified in the previous two tasks. This overlap indicates that, although different metrics were used, they converge toward a shared latent dimension capturing a general notion of research quality or impact. 

The first recurring theme, consistent with the previous analyses, involves \textbf{research fields and field-specific technologies} (e.g., software development, pedagogy). The relationship between these features and venue impact is somewhat weaker than in the earlier task, yet still evident. This indicates that the topical domain of a paper continues to affect the expected impact of its publication venue. Such an effect likely reflects both structural differences among disciplines in citation practices and temporary fluctuations in attention toward specific subfields.

A second prominent theme relates to the \textbf{research methodology} (e.g., 1392, 1618, 8721), which closely aligns with the analogous group of features identified in Task 1. These features capture whether a paper 1) adopts a literature review or survey-based approach, 2) employs a well-defined qualitative methodology, or 3) follows a quantitative approach grounded in standard data analysis techniques.

Their predictive power suggests that LLMs can internalise representations of research methodology and recognise textual cues associated with rigour and standardisation. Papers that adhere to established methodological norms tend to be associated with higher-quality venues, indicating that the model implicitly captures aspects of academic best practices.

Finally, as in the first task, we observe several features linked to \textbf{academic language} (e.g., 6278, 12691, 3864, 15358), encompassing specialized scientific terminology and stylistic conventions. One plausible interpretation is that such linguistic markers signal disciplinary expertise and alignment with the norms of specific research communities, which in turn correlates with publication in more prestigious venues. This resonates with sociolinguistic observations that academic subfields often develop distinct linguistic and methodological identities. However, these linguistic features are the least interpretable, as their meaning can be subtle and context-dependent. Further investigation will be required to clarify how language use interacts with perceived research quality and venue impact.

\section{Conclusions}\label{sec:conclusions}
This paper presents, to the best of our knowledge, the first systematic investigation into how LLMs encapsulate the concept of research quality. We extracted interpretable monosemantic features from LLM activations using sparse autoencoders and analysed their ability to act as predictors across three key scientometric tasks: predicting citation counts, journal SJR, and journal h-index. The results demonstrate that LLMs encode a rich and multidimensional representation of scientific quality, capturing information that aligns with established indicators of research excellence.

Our analysis revealed that the learned features reflect several conceptual dimensions of research quality, including 1) research methodologies, 2) publication type (e.g., survey), 3) high-impact research domains and technologies, and 4) specialised scientific jargon. These dimensions suggest that LLMs not only process linguistic information but also internalise patterns related to the epistemic and structural aspects of scientific communication. We discussed these findings in the broader context of research evaluation and made all experimental materials openly available to facilitate future studies.

This study presents some limitations that we aim to address in future work. First, the interpretation of monosemantic features inherently involves a qualitative component, as semantic labels are assigned by human researchers and may introduce subjectivity. Large-scale studies will therefore be essential to validate and refine these insights. Second, this work focused exclusively on Gemma 2 models, as they were particularly suited for the extraction of monosemantic features; extending the analysis to other architectures will help assess the generality of the results. Third, while this work classified publications into research quality quartiles, future studies could use regression to predict specific values. 
Finally, an intriguing direction for future research concerns how interpretable features can enhance LLM performance in tasks such as research evaluation, academic text generation, and hypothesis discovery, thereby contributing to a deeper understanding of how artificial models capture the structure and quality of scientific knowledge.

\section*{Declaration on Generative AI}  
During the preparation of this work, the authors used ChatGPT: Grammar and spelling check.  After using these tool, the authors reviewed and edited the content as needed and takes full responsibility for the publication’s content.

\bibliography{main}

@article{bolanos2025modelling,
  title={Modelling and Classifying the Components of a Literature Review},
  author={Bola{\~n}os, Francisco and Salatino, Angelo and Osborne, Francesco and Motta, Enrico},
  journal={arXiv preprint arXiv:2508.04337},
  year={2025}
}

@article{salatino2022cso,
  title={Cso classifier 3.0: a scalable unsupervised method for classifying documents in terms of research topics},
  author={Salatino, Angelo and Osborne, Francesco and Motta, Enrico},
  journal={International Journal on Digital Libraries},
  volume={23},
  number={1},
  pages={91--110},
  year={2022},
  publisher={Springer}
}

@inproceedings{john2025scimantify,
  title={SciMantify-a hybrid approach for the evolving semantification of scientific knowledge},
  author={John, Lena and Farfar, Kheir Eddine and Auer, S{\"o}ren and Karras, Oliver},
  booktitle={International Conference on Web Engineering},
  pages={217--225},
  year={2025},
  organization={Springer}
}

@article{dessi2025cs,
  title={CS-KG 2.0: A Large-scale Knowledge Graph of Computer Science},
  author={Dess{\'\i}, Danilo and Osborne, Francesco and Buscaldi, Davide and Reforgiato Recupero, Diego and Motta, Enrico},
  journal={Scientific Data},
  volume={12},
  number={1},
  pages={964},
  year={2025},
  publisher={Nature Publishing Group UK London}
}

@article{osborne2019reducing,
  title={Reducing the effort for systematic reviews in software engineering},
  author={Osborne, Francesco and Muccini, Henry and Lago, Patricia and Motta, Enrico},
  journal={Data Science},
  volume={2},
  number={1-2},
  pages={311--340},
  year={2019},
  publisher={SAGE Publications Sage UK: London, England}
}

@inproceedings{lehmann2024large,
  title={Large language models for scientific question answering: An extensive analysis of the sciqa benchmark},
  author={Lehmann, Jens and Meloni, Antonello and Motta, Enrico and Osborne, Francesco and Recupero, Diego Reforgiato and Salatino, Angelo Antonio and Vahdati, Sahar},
  booktitle={European Semantic Web Conference},
  pages={199--217},
  year={2024},
  organization={Springer}
}

@article{borrego2025research,
  title={Research hypothesis generation over scientific knowledge graphs},
  author={Borrego, Agust{\'\i}n and Dess{\`\i}, Danilo and Ayala, Daniel and Hern{\'a}ndez, Inma and Osborne, Francesco and Recupero, Diego Reforgiato and Buscaldi, Davide and Ruiz, David and Motta, Enrico},
  journal={Knowledge-Based Systems},
  volume={315},
  pages={113280},
  year={2025},
  publisher={Elsevier}
}

@article{aggarwal2026large,
  title={Large language models for scholarly ontology generation: An extensive analysis in the engineering field},
  author={Aggarwal, Tanay and Salatino, Angelo and Osborne, Francesco and Motta, Enrico},
  journal={Information Processing \& Management},
  volume={63},
  number={1},
  pages={104262},
  year={2026},
  publisher={Elsevier}
}

@article{tsaneva2025knowledge,
  title={Knowledge graph validation by integrating LLMs and human-in-the-loop},
  author={Tsaneva, Stefani and Dess{\`\i}, Danilo and Osborne, Francesco and Sabou, Marta},
  journal={Information Processing \& Management},
  volume={62},
  number={5},
  pages={104145},
  year={2025},
  publisher={Elsevier}
}

@article{salatino2025survey,
  title={A survey of knowledge organization systems of research fields: Resources and challenges},
  author={Salatino, Angelo and Aggarwal, Tanay and Mannocci, Andrea and Osborne, Francesco and Motta, Enrico},
  journal={Quantitative Science Studies},
  volume={6},
  pages={567--610},
  year={2025},
  publisher={MIT Press 255 Main Street, 9th Floor, Cambridge, Massachusetts 02142, USA~…}
}

@article{corrin2022importance,
  title={The importance of choosing the right keywords for educational technology publications},
  author={Corrin, Linda and Thompson, Kate and Hwang, Gwo-Jen and Lodge, Jason M},
  journal={Australasian Journal of Educational Technology},
  volume={38},
  number={2},
  pages={1--8},
  year={2022}
}

@inproceedings{devlin2019bert,
  title={Bert: Pre-training of deep bidirectional transformers for language understanding},
  author={Devlin, Jacob and Chang, Ming-Wei and Lee, Kenton and Toutanova, Kristina},
  booktitle={Proceedings of the 2019 conference of the North American chapter of the association for computational linguistics: human language technologies, volume 1 (long and short papers)},
  pages={4171--4186},
  year={2019}
}

@article{xu2020building,
  title={Building a PubMed knowledge graph},
  author={Xu, Jian and Kim, Sunkyu and Song, Min and Jeong, Minbyul and Kim, Donghyeon and Kang, Jaewoo and Rousseau, Justin F and Li, Xin and Xu, Weijia and Torvik, Vetle I and others},
  journal={Scientific data},
  volume={7},
  number={1},
  pages={205},
  year={2020},
  publisher={Nature Publishing Group UK London}
}

@article{cadeddu2024comparative,
  title={A comparative analysis of knowledge injection strategies for large language models in the scholarly domain},
  author={Cadeddu, Andrea and Chessa, Alessandro and De Leo, Vincenzo and Fenu, Gianni and Motta, Enrico and Osborne, Francesco and Recupero, Diego Reforgiato and Salatino, Angelo and Secchi, Luca},
  journal={Engineering Applications of Artificial Intelligence},
  volume={133},
  pages={108166},
  year={2024},
  publisher={Elsevier}
}

@article{laranjo2018conversational,
  title={Conversational agents in healthcare: a systematic review},
  author={Laranjo, Liliana and Dunn, Adam G and Tong, Huong Ly and Kocaballi, Ahmet Baki and Chen, Jessica and Bashir, Rabia and Surian, Didi and Gallego, Blanca and Magrabi, Farah and Lau, Annie YS and others},
  journal={Journal of the American Medical Informatics Association},
  volume={25},
  number={9},
  pages={1248--1258},
  year={2018},
  publisher={Oxford University Press}
}

@article{bolanos2024artificial,
  title={Artificial intelligence for literature reviews: Opportunities and challenges},
  author={Bolanos, Francisco and Salatino, Angelo and Osborne, Francesco and Motta, Enrico},
  journal={Artificial Intelligence Review},
  volume={57},
  number={10},
  pages={259},
  year={2024},
  publisher={Springer}
}

@article{brody2021scite,
  title={Scite},
  author={Brody, Stacy},
  journal={Journal of the Medical Library Association: JMLA},
  volume={109},
  number={4},
  pages={707},
  year={2021},
  publisher={Medical Library Association}
}

@article{buscaldi2024citation,
  title={Citation prediction by leveraging transformers and natural language processing heuristics},
  author={Buscaldi, Davide and Dess{\'\i}, Danilo and Motta, Enrico and Murgia, Marco and Osborne, Francesco and Recupero, Diego Reforgiato},
  journal={Information Processing \& Management},
  volume={61},
  number={1},
  pages={103583},
  year={2024},
  publisher={Elsevier}
}

@article{meloni2023integrating,
  title={Integrating conversational agents and knowledge graphs within the scholarly domain},
  author={Meloni, Antonello and Angioni, Simone and Salatino, Angelo and Osborne, Francesco and Recupero, Diego Reforgiato and Motta, Enrico},
  journal={Ieee Access},
  volume={11},
  pages={22468--22489},
  year={2023},
  publisher={IEEE}
}

@inproceedings{fan2024survey,
  title={A survey on rag meeting llms: Towards retrieval-augmented large language models},
  author={Fan, Wenqi and Ding, Yujuan and Ning, Liangbo and Wang, Shijie and Li, Hengyun and Yin, Dawei and Chua, Tat-Seng and Li, Qing},
  booktitle={Proceedings of the 30th ACM SIGKDD conference on knowledge discovery and data mining},
  pages={6491--6501},
  year={2024}
}

@article{shu2025survey,
  title={A survey on sparse autoencoders: Interpreting the internal mechanisms of large language models},
  author={Shu, Dong and Wu, Xuansheng and Zhao, Haiyan and Rai, Daking and Yao, Ziyu and Liu, Ninghao and Du, Mengnan},
  journal={arXiv preprint arXiv:2503.05613},
  year={2025}
}

@article{paulo2024automatically,
  title={Automatically interpreting millions of features in large language models},
  author={Paulo, Gon{\c{c}}alo and Mallen, Alex and Juang, Caden and Belrose, Nora},
  journal={arXiv preprint arXiv:2410.13928},
  year={2024}
}

@article{andersen2025generative,
  title={Generative Artificial Intelligence (GenAI) in the research process--A survey of researchers’ practices and perceptions},
  author={Andersen, Jens Peter and Degn, Lise and Fishberg, Rachel and Graversen, Ebbe K and Horbach, Serge PJM and Schmidt, Evanthia Kalpazidou and Schneider, Jesper W and S{\o}rensen, Mads P},
  journal={Technology in Society},
  volume={81},
  pages={102813},
  year={2025},
  publisher={Elsevier}
}

@article{cortes2024ai,
  title={AI-assisted prescreening of biomedical research proposals: ethical considerations and the pilot case of “la Caixa” Foundation},
  author={Cort{\'e}s, Carla Carbonell and Parra-Rojas, C{\'e}sar and P{\'e}rez-Lozano, Albert and Arcara, Francesca and Vargas-S{\'a}nchez, Sarasuadi and Fern{\'a}ndez-Montenegro, Raquel and Casado-Mar{\'\i}n, David and Rondelli, Bernardo and L{\'o}pez-Verdeguer, Ignasi},
  journal={Data \& Policy},
  volume={6},
  pages={e49},
  year={2024},
  publisher={Cambridge University Press}
}

@article{thelwall2025evaluating,
  title={Evaluating the predictive capacity of ChatGPT for academic peer review outcomes across multiple platforms},
  author={Thelwall, Mike and Yaghi, Abdallah},
  journal={Scientometrics},
  pages={1--23},
  year={2025},
  publisher={Springer}
}

@article{alkan2025survey,
  title={A Survey on Hypothesis Generation for Scientific Discovery in the Era of Large Language Models},
  author={Alkan, Atilla Kaan and Sourav, Shashwat and Jablonska, Maja and Astarita, Simone and Chakrabarty, Rishabh and Garuda, Nikhil and Khetarpal, Pranav and Pi{\'o}ro, Maciej and Tanoglidis, Dimitrios and Iyer, Kartheik G and others},
  journal={arXiv preprint arXiv:2504.05496},
  year={2025}
}

@article{pirolli1999information,
  title={Information foraging},
  author={Pirolli, Peter and Card, Stuart},
  journal={Psychological review},
  volume={106},
  number={4},
  pages={643},
  year={1999},
  publisher={American Psychological Association}
}

@article{angioni2021aida,
  title={AIDA: A knowledge graph about research dynamics in academia and industry},
  author={Angioni, Simone and Salatino, Angelo and Osborne, Francesco and Recupero, Diego Reforgiato and Motta, Enrico},
  journal={Quantitative Science Studies},
  volume={2},
  number={4},
  pages={1356--1398},
  year={2021},
  publisher={MIT Press One Rogers Street, Cambridge, MA 02142-1209, USA journals-info~…}
}

@String{Computer = "{IEEE} Computer" }

@String{Academic = "Academic Press" }

@String{Springer = "Springer-Verlag" }

@article{bricken2023monosemanticity,
 title={Towards Monosemanticity: Decomposing Language Models With Dictionary Learning},
 author={Bricken, Trenton and Templeton, Adly and Batson, Joshua and Chen, Brian and Jermyn, Adam and Conerly, Tom and Turner, Nick and Anil, Cem and Denison, Carson and Askell, Amanda and Lasenby, Robert and Wu, Yifan and Kravec, Shauna and Schiefer, Nicholas and Maxwell, Tim and Joseph, Nicholas and Hatfield-Dodds, Zac and Tamkin, Alex and Nguyen, Karina and McLean, Brayden and Burke, Josiah E and Hume, Tristan and Carter, Shan and Henighan, Tom and Olah, Christopher},
 year={2023},
 journal={Transformer Circuits Thread},
 note={https://transformer-circuits.pub/2023/monosemantic-features/index.html}
 }

@article{templeton2024scaling,
  title={Scaling Monosemanticity: Extracting Interpretable Features from Claude 3 Sonnet},
  author={Templeton, Adly and Conerly, Tom and Marcus, Jonathan and Lindsey, Jack and Bricken, Trenton and Chen, Brian and Pearce, Adam and Citro, Craig and Ameisen, Emmanuel and Jones, Andy and Cunningham, Hoagy and Turner, Nicholas L and McDougall, Callum and MacDiarmid, Monte and Freeman, C. Daniel and Sumers, Theodore R. and Rees, Edward and Batson, Joshua and Jermyn, Adam and Carter, Shan and Olah, Chris and Henighan, Tom},
  year={2024},
  journal={Transformer Circuits Thread},
 url={https://transformer-circuits.pub/2024/scaling-monosemanticity/index.html}
 }

@article{abramo2011assessing,
  title={Assessing the varying level of impact measurement accuracy as a function of the citation window length},
  author={Abramo, Giovanni and Cicero, Tindaro and D’Angelo, Ciriaco Andrea},
  journal={Journal of Informetrics},
  volume={5},
  number={4},
  pages={659--667},
  year={2011},
  publisher={Elsevier}
}

@article{hodge2011evaluating,
  title={Evaluating journal quality: Is the H-index a better measure than impact factors?},
  author={Hodge, David R and Lacasse, Jeffrey R},
  journal={Research on Social Work Practice},
  volume={21},
  number={2},
  pages={222--230},
  year={2011},
  publisher={Sage Publications Sage CA: Los Angeles, CA}
}

@article{guerrero2012further,
  title={A further step forward in measuring journals’ scientific prestige: The SJR2 indicator},
  author={Guerrero-Bote, Vicente P and Moya-Aneg{\'o}n, F{\'e}lix},
  journal={Journal of informetrics},
  volume={6},
  number={4},
  pages={674--688},
  year={2012},
  publisher={Elsevier}
}

@misc{neuronpedia,
    title = {Neuronpedia: Interactive Reference and Tooling for Analyzing Neural Networks},
    year = {2023},
    note = {Software available from neuronpedia.org},
    url = {https://www.neuronpedia.org},
    author = {Lin, Johnny}
}

@misc{lieberum2024gemmascopeopensparse,
      title={Gemma Scope: Open Sparse Autoencoders Everywhere All At Once on Gemma 2}, 
      author={Tom Lieberum and Senthooran Rajamanoharan and Arthur Conmy and Lewis Smith and Nicolas Sonnerat and Vikrant Varma and János Kramár and Anca Dragan and Rohin Shah and Neel Nanda},
      year={2024},
      eprint={2408.05147},
      archivePrefix={arXiv},
      primaryClass={cs.LG},
      url={https://arxiv.org/abs/2408.05147}, 
}

@misc{gemmateam2024gemma2improvingopen,
      title={Gemma 2: Improving Open Language Models at a Practical Size}, 
      author={Gemma Team and Morgane Riviere and Shreya Pathak and Pier Giuseppe Sessa and Cassidy Hardin and Surya Bhupatiraju and Léonard Hussenot and Thomas Mesnard and Bobak Shahriari and Alexandre Ramé and Johan Ferret and Peter Liu and Pouya Tafti and Abe Friesen and Michelle Casbon and Sabela Ramos and Ravin Kumar and Charline Le Lan and Sammy Jerome and Anton Tsitsulin and Nino Vieillard and Piotr Stanczyk and Sertan Girgin and Nikola Momchev and Matt Hoffman and Shantanu Thakoor and Jean-Bastien Grill and Behnam Neyshabur and Olivier Bachem and Alanna Walton and Aliaksei Severyn and Alicia Parrish and Aliya Ahmad and Allen Hutchison and Alvin Abdagic and Amanda Carl and Amy Shen and Andy Brock and Andy Coenen and Anthony Laforge and Antonia Paterson and Ben Bastian and Bilal Piot and Bo Wu and Brandon Royal and Charlie Chen and Chintu Kumar and Chris Perry and Chris Welty and Christopher A. Choquette-Choo and Danila Sinopalnikov and David Weinberger and Dimple Vijaykumar and Dominika Rogozińska and Dustin Herbison and Elisa Bandy and Emma Wang and Eric Noland and Erica Moreira and Evan Senter and Evgenii Eltyshev and Francesco Visin and Gabriel Rasskin and Gary Wei and Glenn Cameron and Gus Martins and Hadi Hashemi and Hanna Klimczak-Plucińska and Harleen Batra and Harsh Dhand and Ivan Nardini and Jacinda Mein and Jack Zhou and James Svensson and Jeff Stanway and Jetha Chan and Jin Peng Zhou and Joana Carrasqueira and Joana Iljazi and Jocelyn Becker and Joe Fernandez and Joost van Amersfoort and Josh Gordon and Josh Lipschultz and Josh Newlan and Ju-yeong Ji and Kareem Mohamed and Kartikeya Badola and Kat Black and Katie Millican and Keelin McDonell and Kelvin Nguyen and Kiranbir Sodhia and Kish Greene and Lars Lowe Sjoesund and Lauren Usui and Laurent Sifre and Lena Heuermann and Leticia Lago and Lilly McNealus and Livio Baldini Soares and Logan Kilpatrick and Lucas Dixon and Luciano Martins and Machel Reid and Manvinder Singh and Mark Iverson and Martin Görner and Mat Velloso and Mateo Wirth and Matt Davidow and Matt Miller and Matthew Rahtz and Matthew Watson and Meg Risdal and Mehran Kazemi and Michael Moynihan and Ming Zhang and Minsuk Kahng and Minwoo Park and Mofi Rahman and Mohit Khatwani and Natalie Dao and Nenshad Bardoliwalla and Nesh Devanathan and Neta Dumai and Nilay Chauhan and Oscar Wahltinez and Pankil Botarda and Parker Barnes and Paul Barham and Paul Michel and Pengchong Jin and Petko Georgiev and Phil Culliton and Pradeep Kuppala and Ramona Comanescu and Ramona Merhej and Reena Jana and Reza Ardeshir Rokni and Rishabh Agarwal and Ryan Mullins and Samaneh Saadat and Sara Mc Carthy and Sarah Cogan and Sarah Perrin and Sébastien M. R. Arnold and Sebastian Krause and Shengyang Dai and Shruti Garg and Shruti Sheth and Sue Ronstrom and Susan Chan and Timothy Jordan and Ting Yu and Tom Eccles and Tom Hennigan and Tomas Kocisky and Tulsee Doshi and Vihan Jain and Vikas Yadav and Vilobh Meshram and Vishal Dharmadhikari and Warren Barkley and Wei Wei and Wenming Ye and Woohyun Han and Woosuk Kwon and Xiang Xu and Zhe Shen and Zhitao Gong and Zichuan Wei and Victor Cotruta and Phoebe Kirk and Anand Rao and Minh Giang and Ludovic Peran and Tris Warkentin and Eli Collins and Joelle Barral and Zoubin Ghahramani and Raia Hadsell and D. Sculley and Jeanine Banks and Anca Dragan and Slav Petrov and Oriol Vinyals and Jeff Dean and Demis Hassabis and Koray Kavukcuoglu and Clement Farabet and Elena Buchatskaya and Sebastian Borgeaud and Noah Fiedel and Armand Joulin and Kathleen Kenealy and Robert Dadashi and Alek Andreev},
      year={2024},
      eprint={2408.00118},
      archivePrefix={arXiv},
      primaryClass={cs.CL},
      url={https://arxiv.org/abs/2408.00118}, 
}

\section*{Appendix}
Figures~1–9 present the decision tree models trained on the monosemantic features.

\begin{figure}[!h]
    \centering
    \includegraphics[width=1\linewidth]{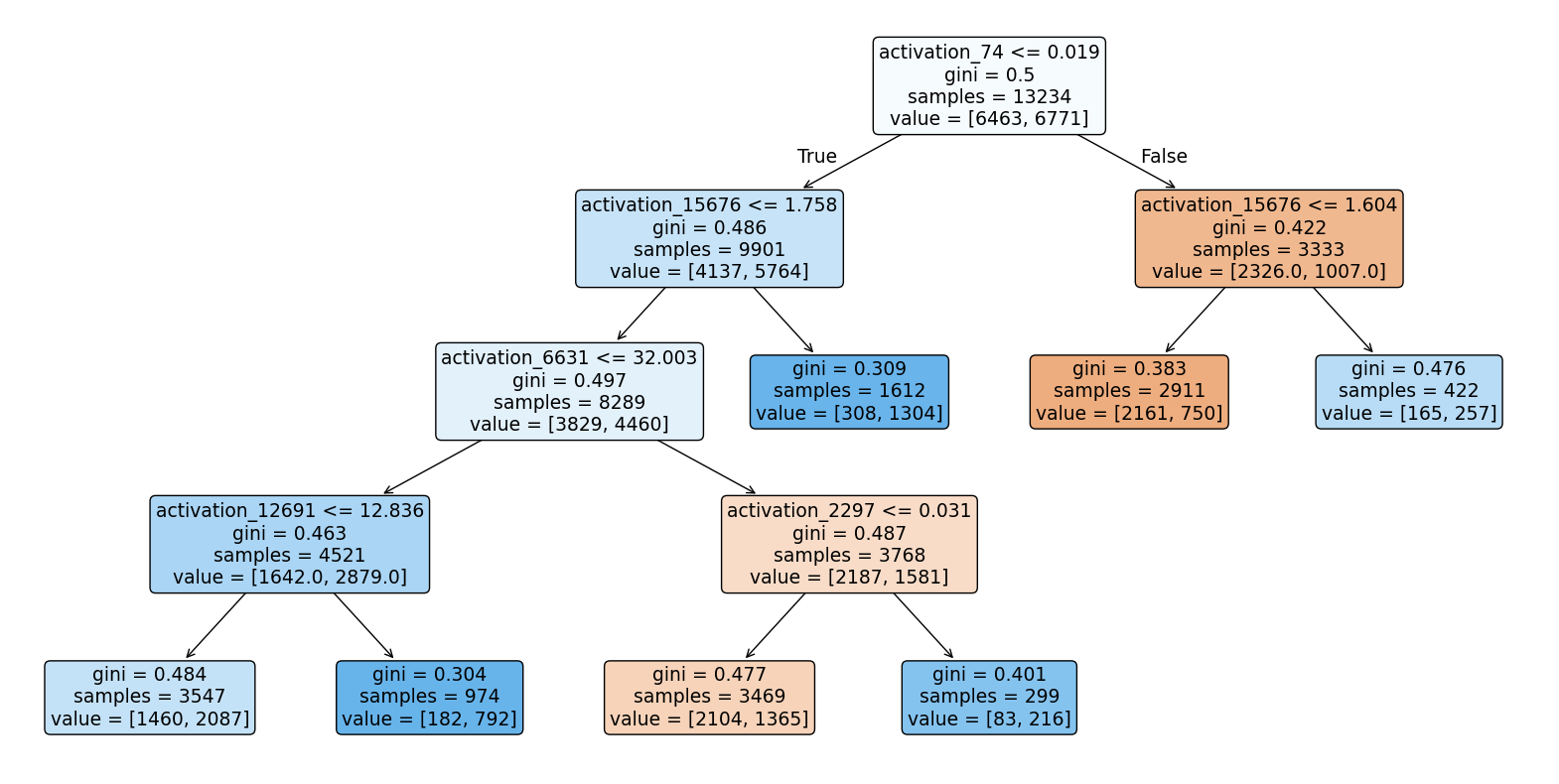}
    \caption{Decision tree obtained from Task 1 using the LLM \& SAE combination 1. Feature definitions are provided in Table \ref{task_1_features}, where \textit{activation\_i} denotes the feature with index \textit{i}.}
    \label{fig:t1_2b_20}
\end{figure}
\begin{figure}[!h]
    \centering
    \includegraphics[width=1\linewidth]{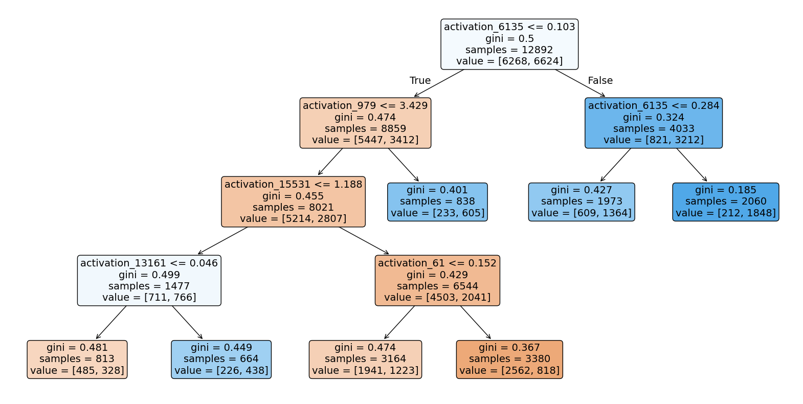}
    \caption{Decision tree obtained from Task 1 using the LLM \& SAE combination 2. Feature definitions are provided in Table \ref{task_1_features}, where \textit{activation\_i} denotes the feature with index \textit{i}.}
    \label{fig:t1_9b_20}
\end{figure}
\begin{figure}[!h]
    \centering
    \includegraphics[width=1\linewidth]{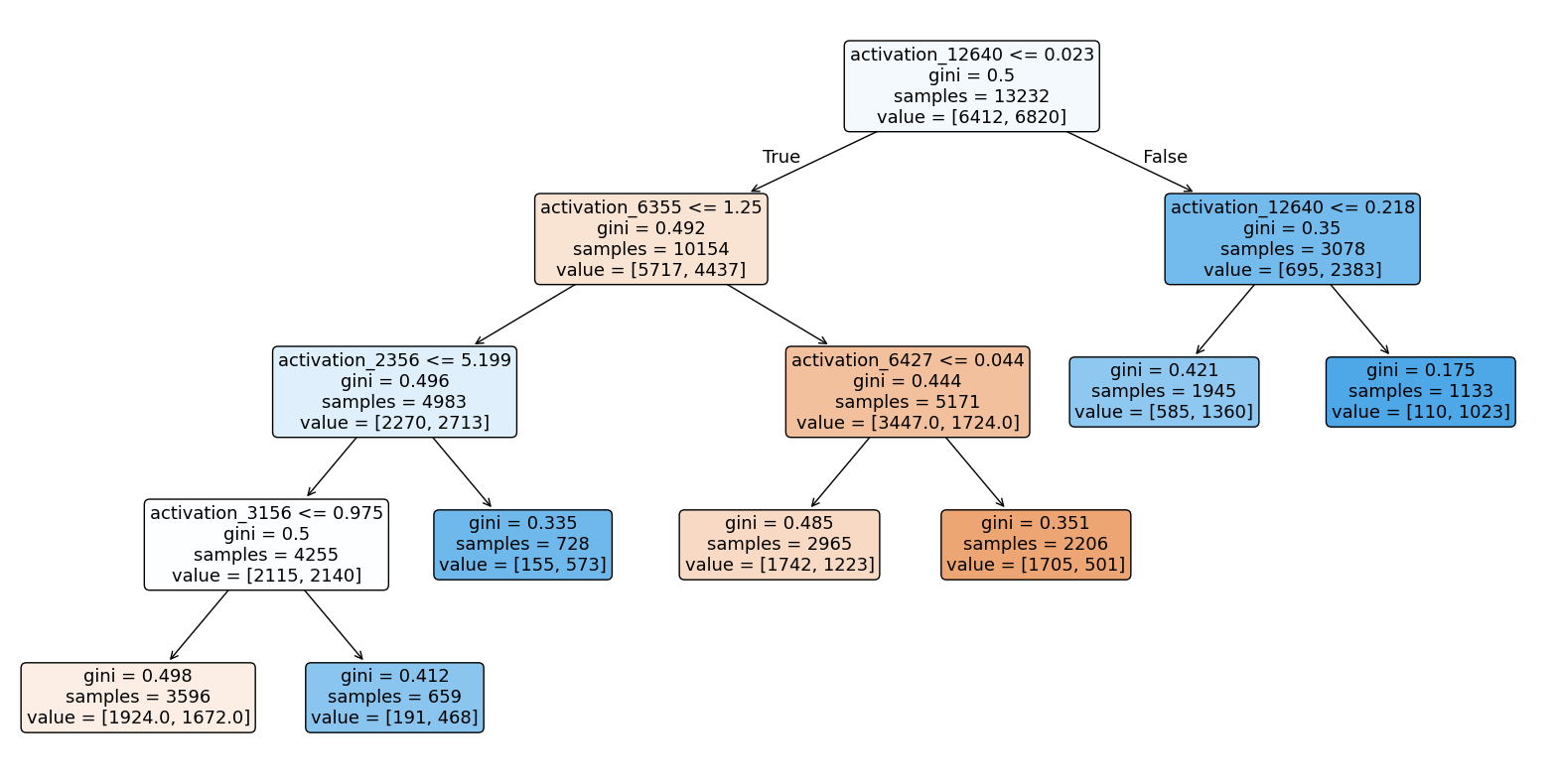}
    \caption{Decision tree obtained from Task 1 using the LLM \& SAE combination 3. Feature definitions are provided in Table \ref{task_1_features}, where \textit{activation\_i} denotes the feature with index \textit{i}.}
    \label{fig:t1_9b_31}
\end{figure}
\begin{figure}[!h]
    \centering
    \includegraphics[width=1\linewidth]{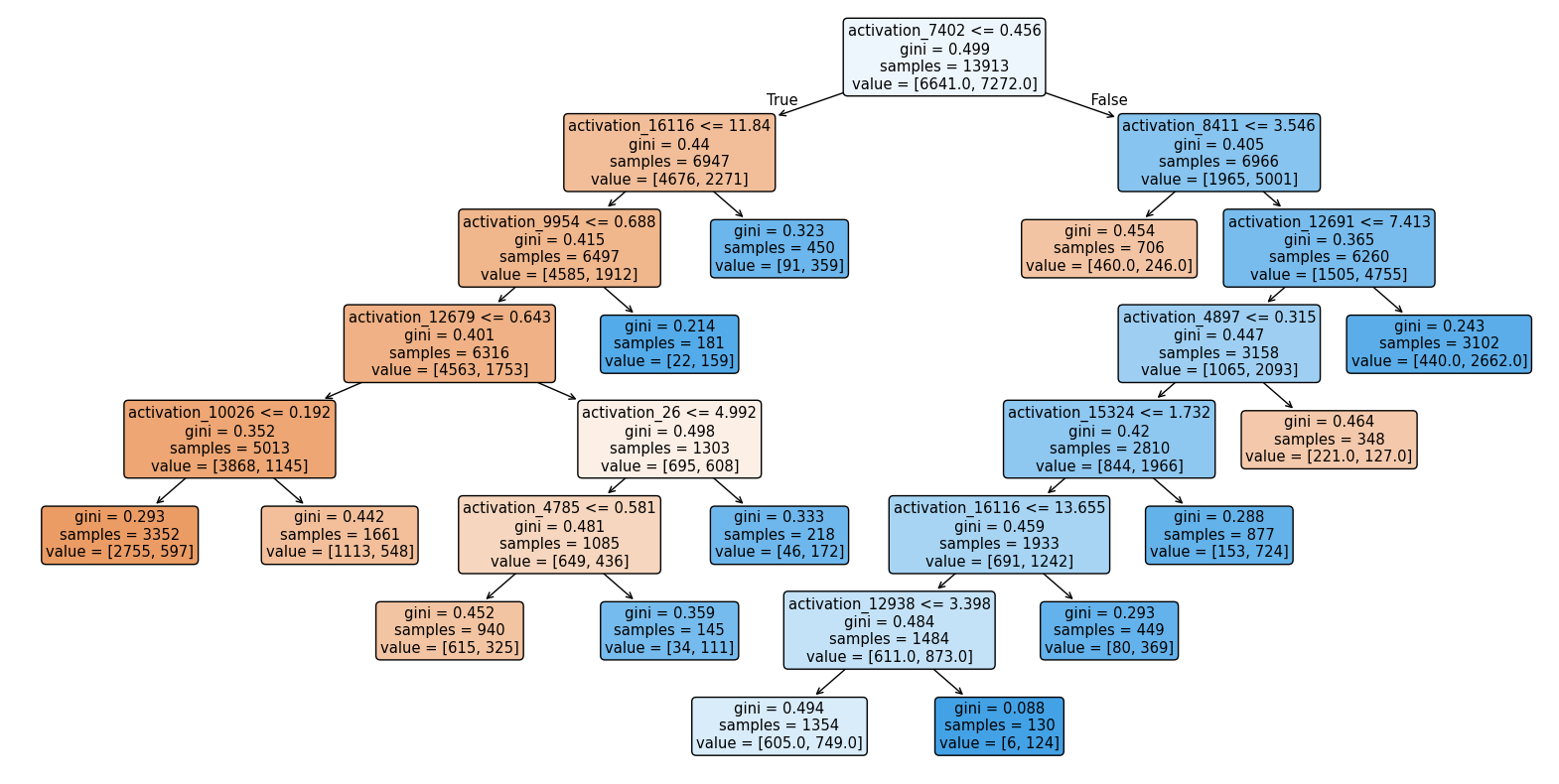}
    \caption{Decision tree obtained from Task 2 using the LLM \& SAE combination 1. Feature definitions are provided in Table \ref{task_2_features}, where \textit{activation\_i} denotes the feature with index \textit{i}.}
    \label{fig:t2_2b_20}
\end{figure}
\begin{figure}[!h]
    \centering
    \includegraphics[width=1\linewidth]{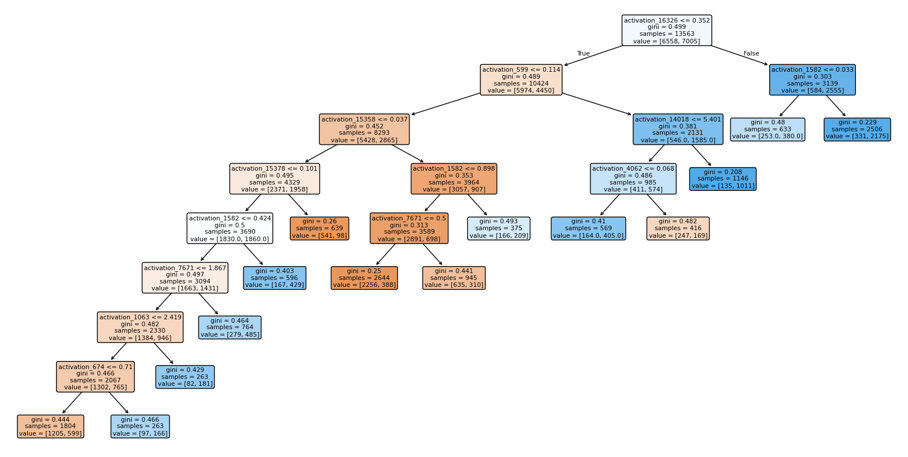}
    \caption{Decision tree obtained from Task 2 using the LLM \& SAE combination 2. Feature definitions are provided in Table \ref{task_2_features}, where \textit{activation\_i} denotes the feature with index \textit{i}.}
    \label{fig:t2_9b_20}
\end{figure}
\begin{figure}[!h]
    \centering
    \includegraphics[width=1\linewidth]{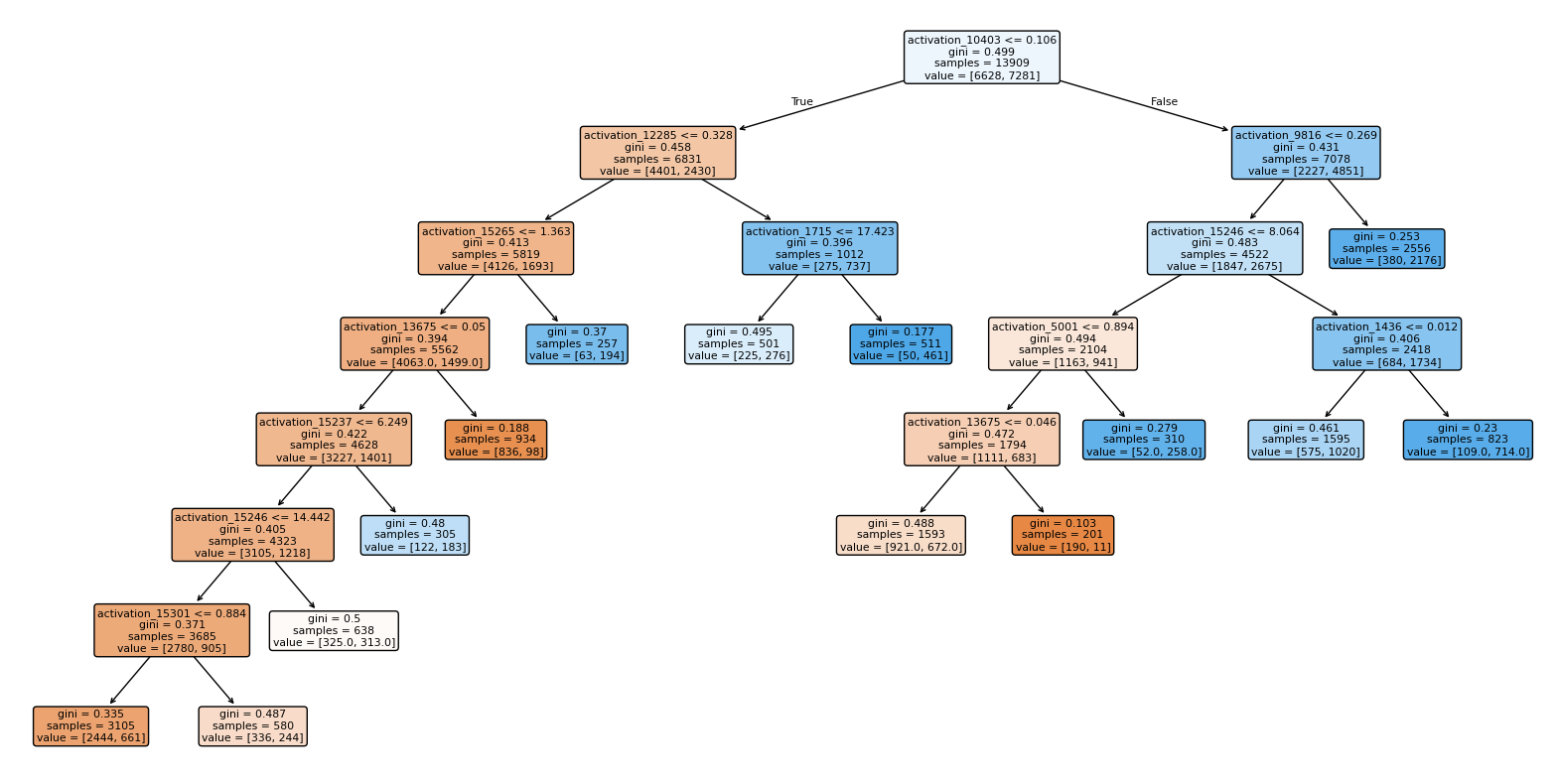}
    \caption{Decision tree obtained from Task 2 using the LLM \& SAE combination 3. Feature definitions are provided in Table \ref{task_2_features}, where \textit{activation\_i} denotes the feature with index \textit{i}.}
    \label{fig:t2_9b_31}
\end{figure}
\begin{figure}[!h]
    \centering
    \includegraphics[width=1\linewidth]{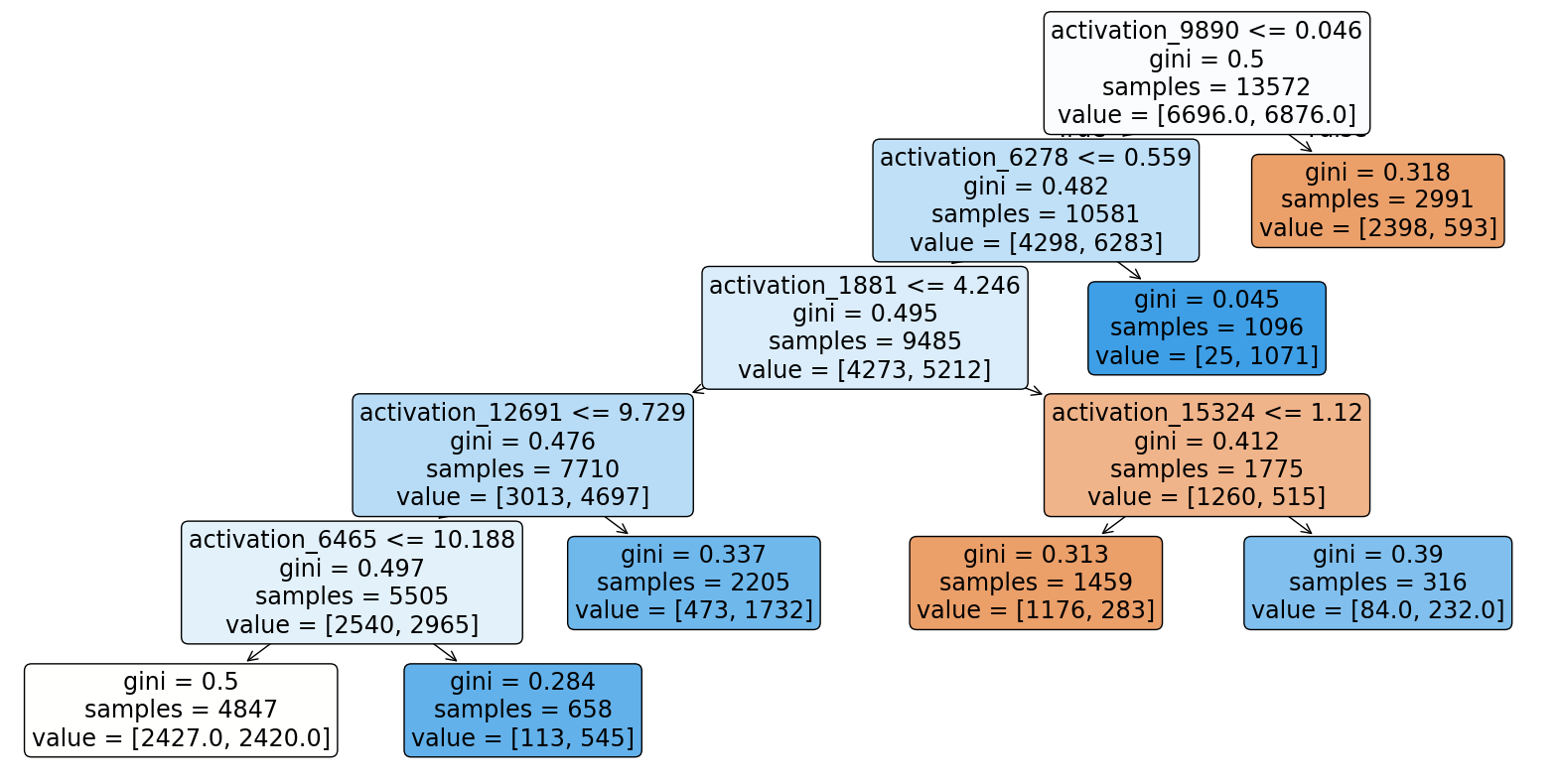}
    \caption{Decision tree obtained from Task 3 using the LLM \& SAE combination 1. Feature definitions are provided in Table \ref{task_3_features}, where \textit{activation\_i} denotes the feature with index \textit{i}.}
    \label{fig:t3_2b_20}
\end{figure}
\begin{figure}[!h]
    \centering
    \includegraphics[width=1\linewidth]{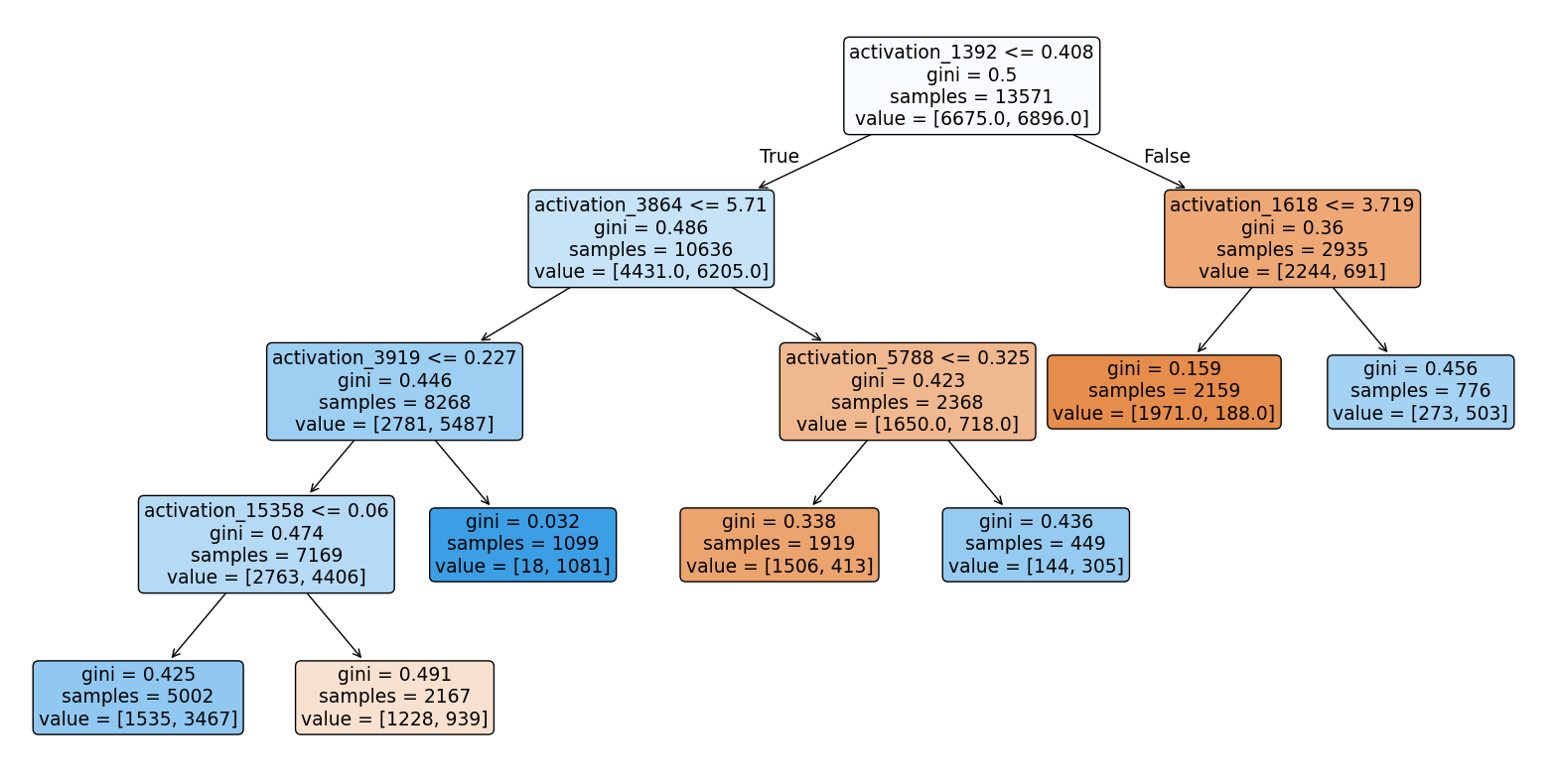}
    \caption{Decision tree obtained from Task 3 using the LLM \& SAE combination 2. Feature definitions are provided in Table \ref{task_3_features}, where \textit{activation\_i} denotes the feature with index \textit{i}.}
    \label{fig:t3_9b_20}
\end{figure}
\begin{figure}[!h]
    \centering
    \includegraphics[width=1\linewidth]{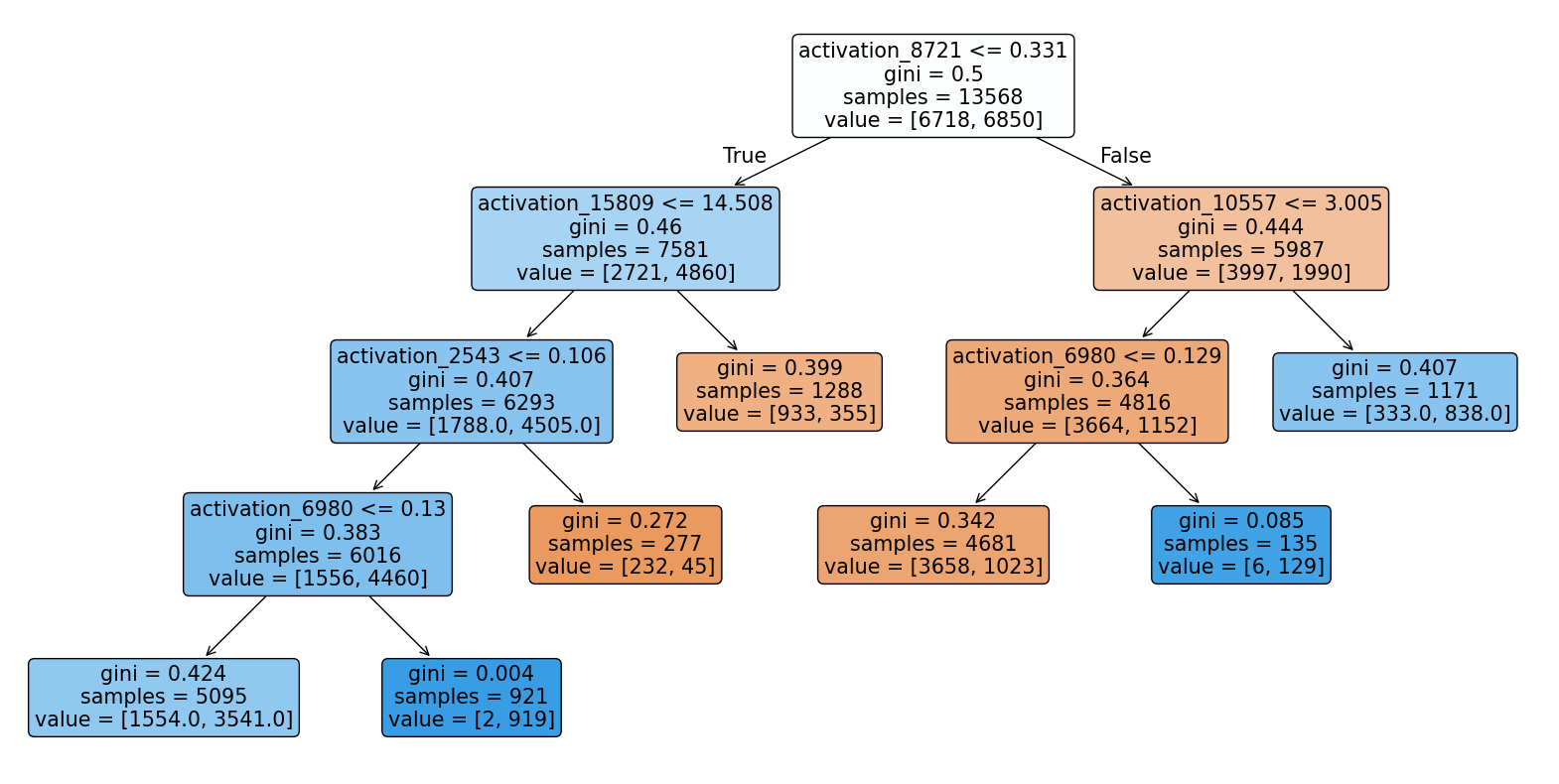}
    \caption{Decision tree obtained from Task 3 using the LLM \& SAE combination 3. Feature definitions are provided in Table \ref{task_3_features}, where \textit{activation\_i} denotes the feature with index \textit{i}.}
    \label{fig:t3_9b_31}
\end{figure}




\end{document}